\documentclass[10pt]{article}

\usepackage[preprint]{tmlr}

\usepackage[T1]{fontenc}
\usepackage{amsmath,amssymb,graphicx,booktabs,array}
\usepackage{tikz}
\usetikzlibrary{arrows.meta,positioning}
\usepackage{placeins}
\usepackage{hyperref}
\usepackage{fontawesome5}

\hypersetup{hidelinks,pdfauthor={},pdftitle={pico-WAM: Evaluating Counterfactual Prediction and Planning Error in a Compact World Action Model}}

\newcommand{\PTdataBlockErrPninetynine}{0.53}

\newcommand{\EvPlanWamBranchRestartExpectedGainPx}{10.9}
\newcommand{\EvPlanWamBranchRestartRealGainPx}{1.2}
\newcommand{\EvCoverageInteraction}{4.8}

\newcommand{\EvCoverageInteractionPos}{12}

\newcommand{\EvCoverageRestartMinusWarmDemo}{22.5}

\newcommand{\EvCoverageRestartMinusWarmBranch}{27.3}

\newcommand{\EvWamDemoRestartCov}{43.8}

\newcommand{\EvWamBranchRestartCov}{78.4}

\newcommand{\EvTestEpisodes}{200}
\newcommand{\EvDevEpisodes}{50}
\newcommand{\EvSteps}{30{,}000}
\newcommand{\EvTrainEpisodes}{164}
\newcommand{\EvValEpisodes}{21}
\newcommand{\EvTestDemoEpisodes}{21}

\newcommand{\MxZeroBlockTest}{0.023}

\newcommand{\MxZeroBlockFresh}{0.027}

\newcommand{\MxRegretDemo}{7.56}
\newcommand{\MxRegretBranch}{2.77}

\newcommand{\MxRuns}{15}
\newcommand{\MxRhoDemo}{0.49}
\newcommand{\MxRhoBranch}{0.80}
\newcommand{\MxStates}{1{,}911}
\newcommand{\MxDemoStates}{470}
\newcommand{\MxExpertStates}{1{,}441}

\newcommand{\FzInteractionMean}{11.8}

\newcommand{\SfOracleRestart}{89.9}

\newcommand{\SfOracleWarm}{49.4}

\newcommand{\ArDetachAuxRestartMean}{\ensuremath{-}44.3}

\newcommand{\FxScoringMean}{12.2}

\newcommand{\SnInteractionLooLow}{4.0}
\newcommand{\SnInteractionLooHigh}{5.6}

\newcommand{\SnScoringLooLow}{11.4}
\newcommand{\SnScoringLooHigh}{12.8}

\newcommand{\nMSE}{\operatorname{nMSE}}

\title{pico-WAM: Evaluating Counterfactual Prediction\\and Planning Error in a Compact World Action Model}
\author{Keivalya Pandya \email pandya.kei@northeastern.edu \\
\addr Khoury College of Computer Sciences\\Northeastern University
}
\begin{document}
\maketitle

\begin{abstract}
When a World Action Model fails at control, is the bottleneck proposal generation or trajectory scoring? We introduce \textbf{pico-WAM}, a compact model with supplied physical state for tracing policy–dynamics interactions. Using 206 Push-T demonstrations and 15 training seeds, we compare data regimes and search configurations. Counterfactual branch training improves short-horizon cost ranking and enlarges the coverage advantage of policy-restarted over warm-started search by 4.8 percentage points; restart wins under both data regimes. Replacing learned trajectory scores with simulator costs improves coverage by 12.2 points while retaining learned proposals. Substituting simulator states in proposal generation alone shows no detected gain. Component diagnostics and a separate door-opening study show why better ranking need not yield better control or outperform imitation. These controlled interventions identify scoring as an effective improvement target in the tested setting and provide a reproducible protocol for diagnosing compact world models.
\end{abstract}

\begin{center}
  \href{https://github.com/keivalya/pico-wam}
       {\faGithub\enspace Code}
\end{center}

\section{Introduction}
pico-WAM is a compact architecture for isolating how a World Action Model predicts outcomes and selects actions. A single shared representation feeds the policy, transition model, and an explicit state readout, making every decision step traceable: observe, propose actions, imagine consequences, score, and execute. We use this transparent setup to study how training data coverage and search choices govern closed-loop behavior.

\begin{figure}[t]
\centering
\includegraphics[width=\linewidth]{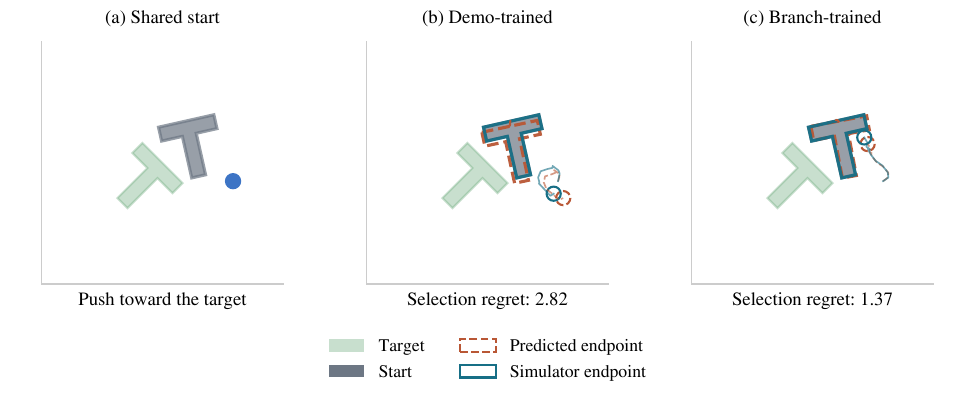}
\caption{\textbf{Comparing candidate choices with their consequences.} Two models select 12-step plans from the same pool of 256 candidates. Wireframe outlines show predicted and simulated physical states rather than decoded images. Neither chosen sequence moves the block; the lower simulator cost here reflects a closer agent approach.}
\label{fig:task}
\end{figure}

A world model trained strictly on demonstrations only observes the actions experts chose, yet a planner must evaluate counterfactual actions they did not choose. Can a compact model score these off-manifold proposals accurately enough to guide control? Figure~\ref{fig:task} illustrates this gap between predicting state transitions and selecting the best action sequence.

We investigate two controlled questions under strictly matched action labels, initial parameters, and training updates. First, does adding counterfactual simulator branches change the benefit of restarting search from the policy versus warm-starting from the previous plan? Second, does replacing learned trajectory scoring with ground-truth simulator costs improve control when retaining learned policy proposals? We find that branch data sharpens cost ranking and favors policy-restarted search, while exact simulator scoring substantially boosts task coverage using the learned proposals. These paired interventions form our core contribution. Component ablations explain the behavior of our shared representation, a supplementary door-opening study extends our ranking findings, and imitation baselines demonstrate why better trajectory ranking alone does not guarantee stronger closed-loop control.

\section{Related Work}
PETS and MBPO address model error through uncertainty-aware search and restricted model rollouts \cite{chua2018pets,janner2019mbpo}. Our added transitions instead come from the true simulator. DINO-WM plans in visual features, PILOT starts from a policy, and SufficientPlan reduces search cost \cite{zhou2025dinowm,liu2026pilot,li2026sufficientplan}. Generative Predictive Control ranks diffusion-policy proposals with a learned model \cite{qi2025gpc}. These establish policy-guided planning; we study its components using supervised state and a known geometric objective.

AD-WM and Action--Consequence Alignment train models for action discrimination \cite{qiu2026adwm,wang2026aca}; AD-WM already measures ranking and selection regret. Decision-relevant error, model exploitation and planning-horizon effects are established concerns \cite{wang2026errors,bhamidipaty2026exploitable,alrasheed2026limits}. DreamTrue expands action coverage using generated videos \cite{li2026dreamtrue}; our simulator supplies paired ground-truth outcomes. Our paired experiments measure how branch data changes the warm/restart gap and how exact scoring helps while retaining learned proposals. These results explain interactions among existing components. Diffusion Policy supplies an imitation reference \cite{chi2023diffusion}; our adapted task does not reproduce published visual-goal benchmark scores.

\section{Model and Planner}
\label{sec:method}
\begin{figure}[t]
\centering
\includegraphics[width=\linewidth]{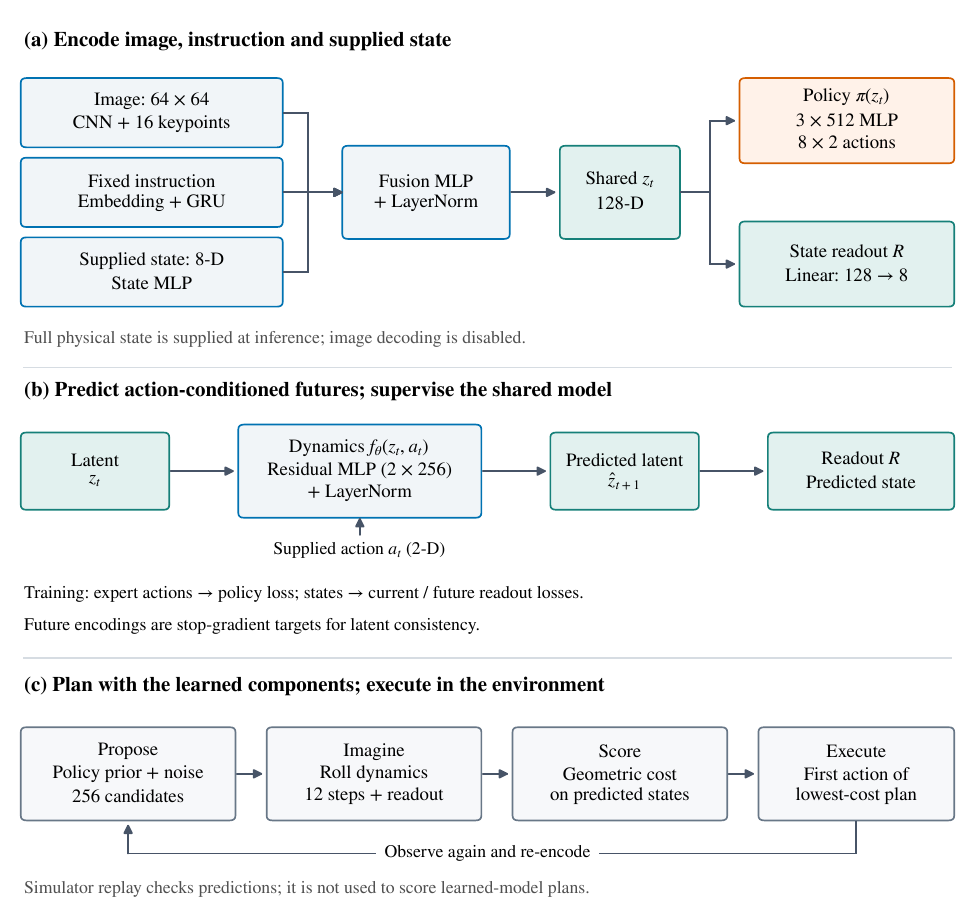}
\caption{\textbf{pico-WAM.} (a) Images and supplied physical state enter a shared latent feeding the policy and readout. Reactive control executes four actions from each eight-action chunk. (b) Actions advance the latent; future latents reuse the same heads. Expert actions, state targets and future encodings supervise the model through Eq.~\ref{eq:objective}. Branches never supply expert action labels. (c) Planning executes one action before observing again. Simulator queries are reserved for diagnostics, except in the explicitly exact planners.}
\label{fig:architecture}
\end{figure}

The policy answers ``what action should I imitate?'' The dynamics answers ``what happens if I take this action?'' A state readout exposes the prediction, and an external planner uses it to compare actions (Fig.~\ref{fig:architecture}).

\subsection{Shared Representation}
Inputs are a $64\times64$ image, a fixed instruction and an eight-dimensional state: agent position/velocity and block position/orientation $(\cos\theta,\sin\theta)$. A three-layer CNN with 16 spatial-softmax keypoints, a GRU and a state MLP feed a fusion MLP, producing $z_t\in\mathbb R^{128}$. The policy has three 512-unit hidden layers and predicts eight two-dimensional actions. Residual dynamics with two 256-unit hidden layers advance the latent:
\begin{equation}
\hat z_{t+1}=f_\theta(z_t,a_t)=\operatorname{LN}\!\left(z_t+\operatorname{MLP}[z_t,a_t]\right).
\end{equation}
A linear readout $R$ maps latents to standardized physical state. Applying the policy to imagined latents supplies an initial plan. The optional image decoder is disabled; full state is available at inference and the instruction is constant. This isolates policy/dynamics interaction rather than visual perception or language generalization.

\subsection{Training}
For a window of $H$ transitions,
\begin{equation}
\mathcal L=\mathcal L_{\rm BC}+0.5(\mathcal L_{s,0}+\mathcal L_{s,1:H})+\mathcal L_z.
\label{eq:objective}
\end{equation}
Each term is squared error averaged over its batch and all output coordinates; future losses also average over $H$. BC regresses eight-action expert chunks. State targets use training-demonstration means and standard deviations, floored at $10^{-3}$. Future latent targets are observation encodings with gradients stopped; prediction gradients reach the dynamics, readout and initial encoder. Action and auxiliary batches are sampled separately (Sec.~4.1). A \emph{branch} perturbs eight demonstration actions from every fourth recorded state with Gaussian noise (SD 0.3), clipped to $[-1,1]^2$. The 6,013 branches yield 48,104 transitions, inherit their source episode's split and supply prediction targets only. Branching changes both state and action coverage.

\subsection{Planning}
The policy rolls forward for 12 steps to form a proposal. The 256 candidates comprise the policy proposal, a search center and 254 noisy perturbations (SD 0.1), clipped to $[-1,1]^2$. We score them using
\begin{equation}
J(\hat s_{1:L})=\frac1L\sum_{k=1}^L\left(w_k d_T(\hat s_k,g)+0.05d_A(\hat s_k)\right).
\label{eq:cost}
\end{equation}
Here $L=12$, $w_k$ increases linearly from 0.5 to 1.5, and $d_T$ averages distances between corresponding block and target corners in pixels. With agent position $p_A$ and block center of mass $c_T$, $d_A=\max(\|p_A-c_T\|_2-15,0)$ subtracts the agent radius. One search iteration selects the cheapest plan's first action before replanning. \emph{Restart} centers search on the current policy proposal (duplicated in the bank); \emph{warm} uses the shifted previous plan, appending the proposal's final action. The first call uses the proposal in both cases.

\section{Experimental Protocol}
\subsection{Data and Matched Training}
Push-T requires pushing a T-shaped block onto a target. The 206 LeRobot demonstrations \cite{lerobotpusht} provide 25,444 transitions at 10\,Hz, with keypoints supplying block pose and controller replay recovering agent velocity. We retain the original geometry and 100\,Hz physics, but use relative actions (up to 100\,px per axis) and rebuild physics each step for exact replay; replayed block transitions agree within \PTdataBlockErrPninetynine\,px at the 99th percentile. Coverage measures maximum target overlap during an episode (including reset), and success requires exceeding 95\%. Because the recordings contain no successes under this threshold (Appendix~\ref{sec:reproduction}), control success requires completion beyond recorded outcomes.

\label{sec:matched}
\emph{BC} learns actions only, \emph{Current} adds current-state reconstruction, and \emph{WAM} optimizes all losses at horizon $H=5$. Current and WAM train on either demonstration-only windows or equal mixtures of demonstrations and counterfactual branches. Within each training run, all five conditions share initial parameters, action minibatches, \EvSteps{} Adam updates, and batch size 64. Demonstration-only conditions resample windows to match auxiliary sample counts, matching updates and labels rather than compute. Lacking trained dynamics, BC and Current act reactively.

Episodes split into \EvTrainEpisodes{} training, \EvValEpisodes{} validation, and \EvTestDemoEpisodes{} test demonstrations. Each of the 15 training runs uses \EvDevEpisodes{} development and \EvTestEpisodes{} final evaluation starts shared across conditions, with warm and restart search sharing checkpoints. Our predeclared primary contrast tests whether branch training increases the advantage of restart over warm search:
\begin{equation}
\Delta=(C^b_{\rm restart}-C^b_{\rm warm})-(C^d_{\rm restart}-C^d_{\rm warm}),
\label{eq:interaction}
\end{equation}
where $C$ is mean coverage and $b,d$ denote branch and demonstration-only training. Four additional coverage contrasts and all success contrasts are secondary. All checkpoints were frozen before final evaluation.

\subsection{Measurements and Interventions}
\label{sec:followups}
Prediction tests roll forward from one encoded observation under recorded or random actions. Normalized mean squared error ($\nMSE$) divides each state-coordinate error by its training standard deviation before squaring and averaging. Random-action tests sample eight sequences in $[-0.4,0.4]^2$ from starts spaced five steps apart, using zero-action outcomes as a reference. Every ten control steps, a non-intervening diagnostic replay compares predicted and true progress of the selected 12-step plan.

For candidate ranking, every model scores the same 256 candidates across \MxStates{} states: \MxDemoStates{} recorded states and \MxExpertStates{} states from 50 fresh noisy-expert episodes (action noise 0.2). Half the candidates are uniform in $[-0.4,0.4]^2$, while half perturb recorded actions (noise 0.1). \emph{Selection regret} is the chosen sequence's true cost minus the best true cost in the pool.

The proposal/scoring study crosses learned (L) and exact (E) components—LL, LE, EL, and EE (proposals first)—reusing 15 frozen branch-trained checkpoints across 256 candidates, a 12-step horizon, and 100 fresh starts per run, with $\mathrm{LE}-\mathrm{LL}$ coverage as primary. Exact scoring evaluates candidate actions in the simulator via Eq.~\ref{eq:cost}, whereas learned scoring uses predicted states. Exact proposals render and re-encode simulator states before querying the policy, replacing the full latent pipeline. Candidate random streams are paired per episode, though closed-loop states naturally diverge. Secondary tests substitute a two-layer, 256-unit state MLP or simulator dynamics while retaining policy and search settings.

Architecture ablations retrain five runs with a single modification: blocked auxiliary encoder gradients, masked image or state inputs (retaining parameter counts), removed instructions, or omitted individual prediction losses. Blocked-gradient minus full-model reactive coverage is primary; gradient blocking tests representation shaping rather than separate encoders.

Additional checks repeat training at $H=12$ with 16-step branches, evaluate frozen models in original \texttt{gym-pusht} 0.1.8 via an action adapter, and compare imitation baselines. Tuned BC sweeps 16 configurations of learning rate, head width, history, and chunk length on separate starts. Diffusion Policy follows its state-based recipe (two observations, 16 predicted/eight executed actions, 100 denoising steps, 5,000 epochs, and EMA weights) across five runs. Comparisons on the dynamics study's starts were selected post-hoc and are exploratory.

\subsection{Door Opening}
A separate study on Meta-World \texttt{door-open-v3} \cite{yu2021metaworld} (39-D state, 4-D actions) trains two-layer, 256-unit state-space dynamics models on 80 scripted-expert episodes alone or mixed equally with 80 noisy-expert episodes (noise 0.3), totaling 5,840 five-step windows per condition. Models share initialization, expert-data normalization, and 10,000 updates at batch size 128, paired with a separate BC policy.

Primary regret is evaluated across 60 states from 20 fresh expert episodes using 64 shared five-step candidates, with cost defined as handle-to-goal distance plus $0.1\times$ hand-to-handle distance. Planners perturb a repeated BC action and execute the best sequence's first step. Secondary control tests use 50 shared starts and 150-step limits across ten runs. Fixed data and evaluation banks mean intervals reflect training-run variance rather than dataset sampling, testing the data-to-ranking mechanism rather than pico-WAM architecture transfer.

\subsection{Uncertainty}
Main-study runs use run-specific start banks paired across conditions, while door-opening runs share a single bank; intervals do not capture dataset or task uncertainty. Secondary intervals are unadjusted, and containing zero does not imply equivalence. Prospective follow-ups use locally timestamped protocols and frozen banks, excluding earlier exploratory runs from confirmatory estimates. Appendix~\ref{sec:reproduction} details reused demonstrations versus fresh evaluation episodes.

\section{What the Components Learn}
\subsection{Broader Data Helps Search}
\label{sec:matched-results}
\begin{figure}[t]
\centering
\includegraphics[width=\columnwidth]{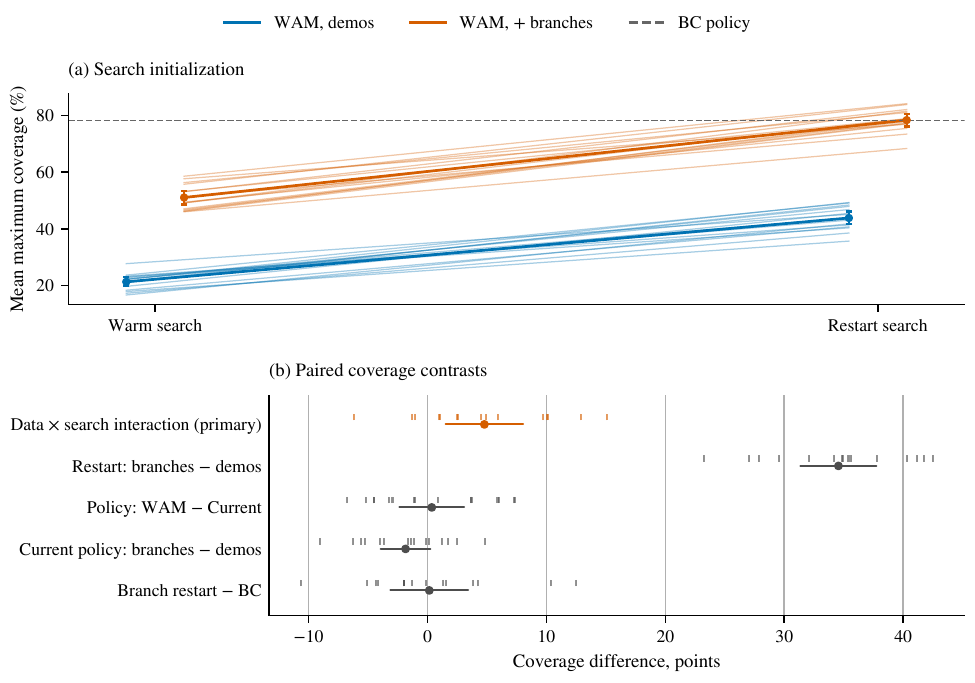}
\caption{\textbf{Branches help restart search more than warm search.} (a) Coverage across shared initial states; thin lines show individual training runs, markers and error bars indicate means and 95\% confidence intervals, and the dashed line denotes BC. (b) Paired coverage contrasts, with ticks marking individual runs. The data-by-search interaction is primary; remaining intervals are secondary and unadjusted.}
\label{fig:evidence}
\end{figure}

\begin{table*}[t]
\caption{Matched study across 15 training runs and 200 initial states per condition. Coverage is reported as mean $\pm$ SD across runs; success is mean percentage. BC and Current act reactively. Per-run counts are provided in the numerical appendix.}
\label{tab:evidence}
\centering\small
\setlength{\tabcolsep}{4pt}
\begin{tabular}{llrr}
\toprule
Training & Control & Coverage, \% & Success, \% \\
\midrule
BC & Reactive & 78.3 $\pm$ 3.3 & 23.3 \\
\addlinespace[2pt]
Current, demos & Reactive & 72.9 $\pm$ 4.2 & 22.2 \\
\addlinespace[2pt]
Current, + branches & Reactive & 71.1 $\pm$ 3.4 & 21.6 \\
\addlinespace[2pt]
WAM, demos & Reactive & 72.1 $\pm$ 2.6 & 22.6 \\
 & Warm search & 21.3 $\pm$ 2.9 & 0.0 \\
 & Restart search & 43.8 $\pm$ 4.1 & 1.8 \\
\addlinespace[2pt]
WAM, + branches & Reactive & 71.4 $\pm$ 2.9 & 20.4 \\
 & Warm search & 51.1 $\pm$ 4.4 & 5.6 \\
 & Restart search & 78.4 $\pm$ 4.1 & 32.8 \\
\bottomrule
\end{tabular}

\end{table*}

Adding counterfactual branches raises policy-restarted search coverage from \EvWamDemoRestartCov\% to \EvWamBranchRestartCov\% (Table~\ref{tab:evidence}). While restart search already outperforms warm search when trained on demonstrations alone, branch training widens this advantage from \EvCoverageRestartMinusWarmDemo{} to \EvCoverageRestartMinusWarmBranch{} percentage points—yielding a primary interaction of \EvCoverageInteraction{} points that is positive in \EvCoverageInteractionPos{} of 15 runs (Fig.~\ref{fig:evidence}). Without branch data, both planners underperform the same checkpoint's reactive policy. Overall, these results favor policy-restarted search under both data regimes rather than reversing the preferred initialization strategy.

\begin{figure}[t]
\centering
\includegraphics[width=\linewidth]{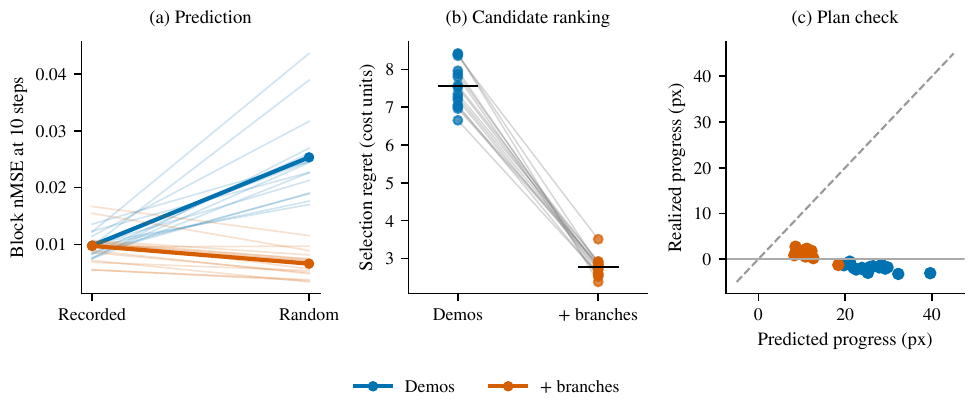}
\caption{\textbf{Prediction, ranking, and control reveal distinct error modes.} (a) Ten-step block prediction error; thin lines show individual runs and thick lines denote means. (b) Selection regret across 1,911 shared states; paired lines denote runs and black markers indicate means. (c) Predicted versus realized plan progress per run, with the dashed line marking exact agreement. Panels evaluate distinct state populations and do not establish causal mediation.}
\label{fig:mechanism}
\end{figure}

\begin{table}[t]
\caption{Prediction and ranking diagnostics: means across \MxRuns{} training runs and paired differences (+ branches minus demos) with 95\% intervals. $^a$Zero-action reference: \MxZeroBlockTest{} on test demonstrations, \MxZeroBlockFresh{} on fresh episodes. $^b$Cost units from Eq.~(\ref{eq:cost}), approximately in pixels; regret is the selected sequence's true cost minus the best candidate's true cost, and optimism is true minus predicted cost. $^c$Evaluated on each model's own restart candidates.}
\label{tab:mechanism}
\centering\small
\setlength{\tabcolsep}{5pt}
\begin{tabular}{@{}lrrr@{}}
\toprule
Measure & Demos & + branches & Difference \\
\midrule
\multicolumn{4}{@{}l}{\emph{Prediction, 21 unseen test demonstrations}} \\
State nMSE, recorded, 15 steps & 0.011 & 0.012 & 0.000 [\ensuremath{-}0.001, 0.002] \\
Block nMSE, random, 10 steps$^a$ & 0.025 & 0.007 & \ensuremath{-}0.019 [\ensuremath{-}0.023, \ensuremath{-}0.015] \\
Block RMSE, px, random, 10 steps & 14.8 & 7.1 & \ensuremath{-}7.8 [\ensuremath{-}9.3, \ensuremath{-}6.2] \\
\multicolumn{4}{@{}l}{\emph{Prediction, 50 fresh scripted-expert episodes}} \\
State nMSE, recorded, 15 steps & 0.042 & 0.020 & \ensuremath{-}0.022 [\ensuremath{-}0.030, \ensuremath{-}0.014] \\
Block nMSE, random, 10 steps$^a$ & 0.041 & 0.015 & \ensuremath{-}0.026 [\ensuremath{-}0.032, \ensuremath{-}0.020] \\
Block RMSE, px, random, 10 steps & 17.3 & 10.6 & \ensuremath{-}6.7 [\ensuremath{-}8.5, \ensuremath{-}4.9] \\
\multicolumn{4}{@{}l}{\emph{Candidate ranking, 1{,}911 common states}} \\
Rank correlation with true cost & 0.49 & 0.80 & 0.32 [0.30, 0.34] \\
Overlap of top 32 & 0.40 & 0.65 & 0.26 [0.24, 0.27] \\
Selection regret$^b$ & 7.56 & 2.77 & \ensuremath{-}4.79 [\ensuremath{-}5.11, \ensuremath{-}4.47] \\
Selection optimism$^b$ & 5.93 & 2.27 & \ensuremath{-}3.66 [\ensuremath{-}4.72, \ensuremath{-}2.60] \\
Gain over policy prior$^{b,c}$ & 4.67 & 7.01 & 2.34 [2.20, 2.47] \\
\bottomrule
\end{tabular}

\end{table}

On held-out test demonstrations, prediction errors under recorded expert actions are comparable across models. Under random actions, however, the demonstration-only model's error exceeds the zero-action baseline on average, whereas branch training reduces block prediction error roughly threefold. This advantage persists on fresh noisy-expert episodes (Table~\ref{tab:mechanism}).

Across shared evaluation states, branch training increases candidate rank correlation from \MxRhoDemo{} to \MxRhoBranch{} and lowers selection regret from \MxRegretDemo{} to \MxRegretBranch{}, improving all 15 runs. Even so, selected plans remain systematically optimistic: branch-trained restart search predicts \EvPlanWamBranchRestartExpectedGainPx\,px of progress but achieves only \EvPlanWamBranchRestartRealGainPx\,px. Because regret evaluates Eq.~\ref{eq:cost}—which combines block-to-goal distance with an agent-approach penalty—this aggregate improvement does not separate approach gains from actual block manipulation, underscoring that ranking a fixed candidate pool differs from completing a full episode.

\subsection{Separate Proposals from Scores}
\begin{figure}[t]
\centering
\includegraphics[width=\linewidth]{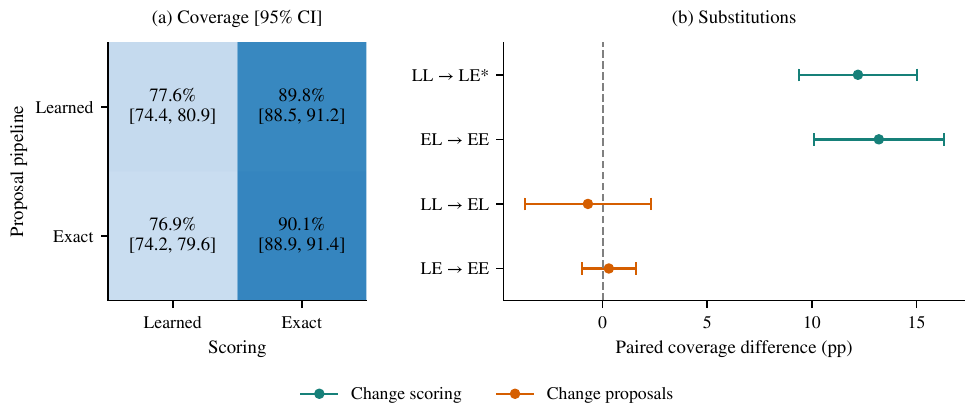}
\caption{\textbf{Exact scoring improves control even with learned proposals retained.} Evaluated across 15 frozen checkpoints on 100 fresh starts each. L and E denote learned and exact computations (proposals listed first). (a) Mean coverage and 95\% intervals. (b) Paired differences; * marks the primary contrast, while others are secondary and unadjusted. Exact proposals include simulator rendering and re-encoding.}
\label{fig:factorial}
\end{figure}

Replacing the entire learned model with the ground-truth simulator raises restart coverage to \SfOracleRestart\%, but simultaneously alters both proposal generation and trajectory scoring. Our crossed factorial evaluation disentangles the two (Fig.~\ref{fig:factorial}): pairing exact simulator scoring with learned policy proposals boosts coverage by \FxScoringMean{} points, with positive gains in every run. Conversely, supplying exact proposals under learned scoring yields no detectable improvement. Exact scoring improves coverage while retaining learned proposals; substituting simulator states during proposal generation alone shows no detected gain. This identifies scoring as an effective intervention in the tested planner.

Notably, even exact simulator dynamics leave warm search at \SfOracleWarm\% coverage, failing to close the gap with policy-restarted search; the relative roles of planning horizon, cost shaping, and search breadth remain open questions. Moreover, a separately trained state-space MLP shows no detectable coverage difference from the shared latent dynamics model while achieving a higher binary success rate (Table~\ref{tab:replication}), confirming that a shared representation is a diagnostic design choice rather than a guaranteed performance advantage.

\subsection{Which Architectural Components Matter?}
\begin{table}[t]
\caption{Ablation variants minus the full model: coverage differences with paired 95\% intervals across five runs. $^a$Reactive coverage with blocked auxiliary gradients is primary; other contrasts are secondary and unadjusted.}
\label{tab:architecture}
\centering\small
\setlength{\tabcolsep}{5pt}
\begin{tabular}{@{}lrr@{}}
\toprule
Variant minus full & Reactive & Restart \\
\midrule
Auxiliary gradients blocked$^a$ & 5.1 [\ensuremath{-}5.1, 15.3] & \ensuremath{-}44.3 [\ensuremath{-}50.2, \ensuremath{-}38.4] \\
Image only & \ensuremath{-}31.5 [\ensuremath{-}37.9, \ensuremath{-}25.1] & \ensuremath{-}41.6 [\ensuremath{-}46.7, \ensuremath{-}36.4] \\
State only & 0.7 [\ensuremath{-}8.0, 9.4] & \ensuremath{-}3.3 [\ensuremath{-}7.1, 0.5] \\
No instruction & 1.5 [\ensuremath{-}6.3, 9.3] & \ensuremath{-}4.1 [\ensuremath{-}8.6, 0.5] \\
No future-state loss & 0.4 [\ensuremath{-}5.2, 5.9] & \ensuremath{-}6.4 [\ensuremath{-}16.6, 3.8] \\
No latent loss & 0.5 [\ensuremath{-}8.2, 9.2] & \ensuremath{-}5.2 [\ensuremath{-}14.3, 3.9] \\
\midrule
Full model, coverage \% & 71.6 & 81.7 \\
\bottomrule
\end{tabular}

\end{table}

Stopping prediction-loss gradients into the encoder leaves the primary reactive-policy contrast inconclusive, but alters policy-restarted search coverage by \ArDetachAuxRestartMean{} points in our secondary analysis (Table~\ref{tab:architecture}); representations shaped solely by behavior cloning do not support equally effective search. Masking physical state inputs and relying on images alone sharply degrades both reactive and planned control, confirming that our compact implementation depends heavily on explicit state. Omitting remaining auxiliary losses or instruction inputs yields no detectable change, though confidence intervals remain too wide to prove they are redundant. Comparing WAM against Current similarly shows no detectable reactive gain from jointly adding multi-step state and latent-consistency losses.

\subsection{Better Choices Need Not Beat Imitation}
\begin{figure}[!ht]
\centering
\includegraphics[width=\linewidth]{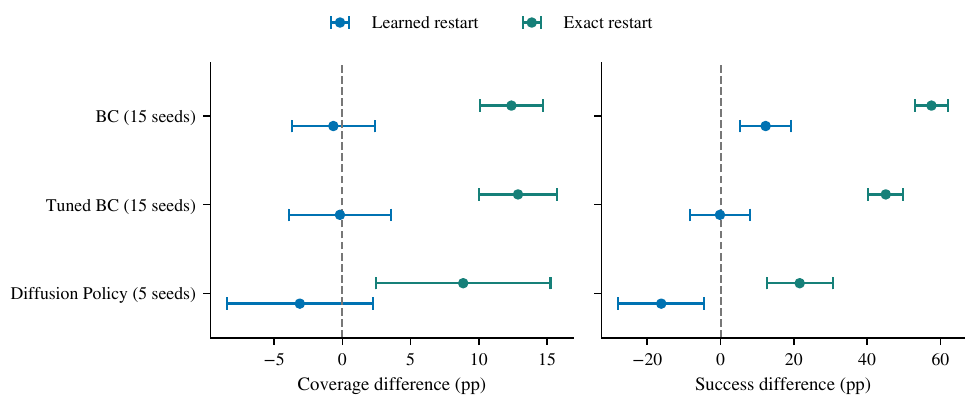}
\caption{\textbf{Imitation baselines separate better trajectory ranking from better closed-loop control.} Exploratory planner-minus-imitation contrasts with unadjusted paired 95\% intervals; positive values favor planning. Comparisons share training runs and initial states (five runs for Diffusion Policy). Exact restart is a diagnostic that changes both proposal generation and scoring.}
\label{fig:imitation}
\end{figure}

\begin{table}[!ht]
\caption{Branch-trained restart search minus imitation baselines (top), and alternative dynamics minus latent dynamics (bottom), reported in percentage points with paired 95\% intervals. Tuned-BC and Diffusion Policy rows use imitation-study starts, whereas Fig.~\ref{fig:imitation} uses dynamics-study starts. Success contrasts are secondary.}
\label{tab:replication}
\centering\small
\setlength{\tabcolsep}{5pt}
\begin{tabular}{@{}llrr@{}}
\toprule
Study, seeds & Reference & Coverage & Success \\
\midrule
Matched, 15 & BC & 0.1 [\ensuremath{-}3.1, 3.4] & 9.5 [3.4, 15.6] \\
Horizon-matched, 15 & BC & 0.2 [\ensuremath{-}4.0, 4.4] & 10.0 [4.7, 15.3] \\
Original gym-pusht, 15 & BC & \ensuremath{-}1.4 [\ensuremath{-}4.6, 1.7] & 8.1 [1.7, 14.5] \\
Tuned BC, 15 & tuned BC & 2.7 [\ensuremath{-}0.6, 6.0] & \ensuremath{-}0.1 [\ensuremath{-}7.3, 7.1] \\
Diffusion Policy, 5 & DP & \ensuremath{-}1.0 [\ensuremath{-}7.6, 5.5] & \ensuremath{-}16.7 [\ensuremath{-}27.9, \ensuremath{-}5.5] \\
\midrule
State dynamics, 15 & latent & \ensuremath{-}1.9 [\ensuremath{-}4.7, 1.0] & 12.0 [5.7, 18.3] \\
Oracle dynamics, 15 & latent & 13.1 [10.6, 15.5] & 45.3 [38.5, 52.0] \\
\bottomrule
\end{tabular}

\end{table}

Training with a matched 12-step horizon preserves a positive data-by-search interaction (\FzInteractionMean{} points). Evaluating in the original \texttt{gym-pusht} (v0.1.8) environment confirms the individual benefits of branch data and policy-restarted search, though omitting demonstration-only warm search precludes testing the full interaction. Across both checks, branch-trained restart search shows no detectable coverage advantage over base BC (Table~\ref{tab:replication}). In our matched study, restart search improves binary success over untuned BC, but shows no detectable advantage over tuned BC and trails Diffusion Policy in overall success rate (Fig.~\ref{fig:imitation}). Because reactive policies execute four-action chunks between observations while planners replan at every step, comparing them conflates search with feedback frequency; by contrast, our primary warm/restart and oracle-scoring contrasts hold replanning frequency constant.

\begin{figure}[!ht]
\centering
\includegraphics[width=\linewidth]{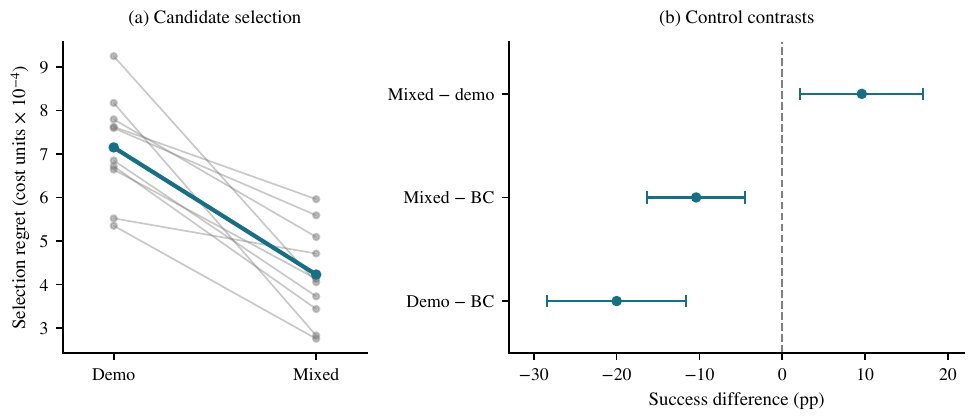}
\caption{\textbf{Broader transition data also improves cost ranking in door opening.} Evaluated across ten training runs with fixed datasets and candidate banks. (a) Paired selection regret, highlighting the mean. (b) Secondary success rate differences with unadjusted paired 95\% intervals. Standalone state-space models test the data-to-ranking mechanism rather than shared-representation transfer.}
\label{fig:contact}
\end{figure}

In Meta-World door opening, augmenting demonstrations with noisy rollouts reduces selection regret across all ten runs, dropping from $7.15\times10^{-4}$ to $4.23\times10^{-4}$ cost units (paired difference $-2.92\times10^{-4}$, 95\% CI $[-3.94,-1.90]\times10^{-4}$). Planning with mixed-data dynamics succeeds in 88.2\% of trials compared to 78.6\% for demonstration-only dynamics, though both trail the reactive BC policy at 98.6\% (Fig.~\ref{fig:contact}). While cost units are not directly comparable to Push-T, adding noisy-expert rollouts improves candidate ranking and planner success in this door-opening study, while both planners remain below reactive BC in success rate.

\section{Discussion and Conclusion}
pico-WAM makes a WAM's computation visible: one representation supports imitation, imagined transitions and state prediction. The experiments turn that architecture into practical lessons. Test predictions on actions the planner might choose, rather than only recorded actions. Separate proposal generation from scoring when diagnosing a plan. Measure the data-by-search interaction and retain a direct-action control.

Branches enlarge an existing restart advantage; they do not change which initialization wins. The scoring substitution identifies an effective intervention conditional on learned proposals. Lower regret can reflect agent approach before any block movement (Fig.~\ref{fig:task}), so it need not imply immediate target progress. Branch-trained plans still overestimate realized progress; identifying which contact or state errors drive each failure remains open. These interventions explain behavior in a controlled setting; they do not establish a new architecture, hardware transfer, or general superiority of planning. The shared-model tests use supervised state in Push-T, and the door experiment uses a different architecture. This scope lets a small teaching implementation support reproducible questions about how acting and predicting interact, without requiring it to compete with large models.

\FloatBarrier
\clearpage
\bibliographystyle{tmlr}
\bibliography{references}
\clearpage
\appendix
\section{Reproduction Details}
\label{sec:reproduction}
\paragraph{Study history and scope}
A five-seed development study triggered a predeclared precision rule; its final bank was never evaluated. The 15-seed study used new final starts and frozen checkpoints. Test demonstrations had appeared in earlier exploration; fresh control starts and prediction episodes provide separate checks. Reports retain protocols, hashes and per-run outcomes.

\paragraph{Retrospective sensitivity}
Omitting one run at a time gives primary interaction estimates of \SnInteractionLooLow--\SnInteractionLooHigh{} points and scoring effects of \SnScoringLooLow--\SnScoringLooHigh{}. All corresponding paired-$t$ intervals remain above zero. The supplement also reports 50,000 paired-run bootstrap draws. These checks reuse the original runs; they do not extend dataset/task generality or replace the primary analyses.

\paragraph{Executable supplement}
\texttt{reproduction-review.zip} contains source, configurations, archived reports and pinned review dependencies. A fresh Python 3.11 CPU environment regenerated all eight figures and matching numerical files, replayed the saved Fig.~\ref{fig:task} actions, and passed a miniature factorial pipeline. Full retraining, checkpoints and original data downloads are separate; commands and upstream revisions are included. These checks do not establish full numerical reproduction.

\paragraph{Resources}
On a V100-SXM2, 30,000 updates took 195\,s for BC and 465\,s for WAM; restart control took about 13\,ms/step including diagnostic replay. Main-study work used 21.3 GPU-hours, horizon-matched training 16.0, architecture tests 16.9 and Diffusion Policy 44.7. The dynamics study used 1,390 allocated CPU-hours, including reruns. These are implementation-specific measurements, not an efficiency comparison or a complete compute total.

\begin{figure}[!h]
\centering
\includegraphics[width=.75\linewidth]{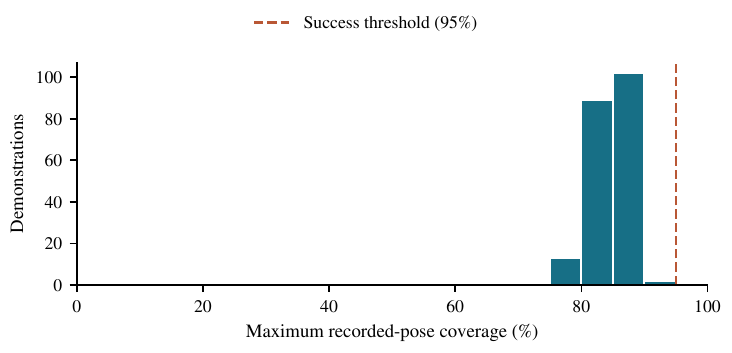}
\caption{Recorded demonstrations remain below the 95\% success threshold (mean maximum coverage 84.8\%). Original gym-pusht and reconstructed coverage agree within $1.7\times10^{-14}$. Source rewards and success flags confirm that incomplete demonstrations precede conversion.}
\label{fig:demo-coverage}
\end{figure}

\paragraph{Data checks}
All 25,650 source frames agree across the demonstration and keypoint repositories in indices, rewards and flags. Reconstructed corners agree within $3.5\times10^{-5}$\,px; next-state coverage reproduces rewards within $2.7\times10^{-7}$ across 25,444 transitions. No recorded frame succeeds (Fig.~\ref{fig:demo-coverage}); the human recording termination policy is unknown. The study therefore concerns learning from incomplete demonstrations.

\paragraph{Door implementation}
Meta-World source tag v3.0.0 is paired with MuJoCo 3.6.0; its package metadata reports 2.0.0. Restored simulator continuations passed the registered $10^{-6}$ tolerance. Fifty final starts are shared across ten seeds, not 500 unique starts. Source hashes and package versions are retained in the supplementary reports. Simulator branching requires arbitrary-state replay and is not a hardware collection procedure.

\end{document}